\documentclass[12pt]{article}
\usepackage{amsmath, amssymb, amsthm, graphicx, geometry}
\usepackage[authoryear]{natbib}
\usepackage[ruled, linesnumbered, vlined]{algorithm2e}
\usepackage{hyperref}
\usepackage[utf8]{inputenc} 
\usepackage[T1]{fontenc}    
\usepackage{url}            
\usepackage{booktabs}       
\usepackage{amsfonts}       
\usepackage{nicefrac}       
\usepackage{microtype}      
\usepackage{xcolor}         
\usepackage{stmaryrd}
\usepackage{multirow}
\usepackage{graphicx}
\usepackage{xcolor}
\usepackage{mathrsfs,mathtools}
\usepackage{bm, bbm}
\usepackage{authblk}
\usepackage{listings}

\title{NuclearQAv2: A Structured Benchmark\\ for Evaluating Domain-Science Competence\\ in Large Language Models}
\author{Henry Shaowu Yuchi, Michal Kucer, Benjamin H. Sims\\
Selma Peterson, Emily Taylor\\
\vspace{1em}
Los Alamos National Laboratory
}
\date{}

\begin{document}

\maketitle

\begin{abstract}
Large language models (LLMs) have demonstrated strong performance across a wide range of tasks, but ensuring their reliability in highly technical domains remains a significant challenge. In nuclear engineering, problem solving often requires not only factual knowledge but also quantitative reasoning and conceptual understanding.
To address the need for systematic evaluation in this domain, we introduce NuclearQAv2, a benchmark for assessing LLMs on nuclear engineering knowledge. The benchmark comprises approximately 1,240 question-answer pairs spanning three categories: boolean, numeric, and verbal. NuclearQAv2 is constructed using a hybrid pipeline that combines expert-authored questions, existing datasets, and LLM-assisted generation from domain-specific technical corpora. By leveraging structured prompting for both automated question generation and response evaluation, the proposed framework enables scalable benchmark construction and evaluation.
We evaluate a diverse set of LLMs using NuclearQAv2 and observe substantial performance differences across task types. While the models generally perform well on factual questions, quantitative reasoning and conceptual understanding remain considerably more challenging. These results highlight the importance of multi-faceted evaluation frameworks and establish NuclearQAv2 as a scalable benchmark for assessing LLM capabilities in technical domains.

\end{abstract}

\noindent%
{\it Keywords:} Large language model, benchmark, nuclear engineering.
\vfill

\newpage

\section{Introduction}
\label{sec:intro}


Despite their rapid rise in popularity, large language models (LLMs) have demonstrated strong performance across a wide range of language understanding and knowledge-intensive tasks, including translation, question answering, multi-task reasoning, and text generation \citep{brown2020language, hendrycks2020measuring, achiam2023gpt, team2023gemini}. Consequently, they are being investigated and deployed in many technical domains with high impact, such as medicine, law, finance, and education \citep{singhal2023large,chen2024survey,wang2026large}. 
However, LLMs are known to exhibit \emph{hallucination}, generating plausible yet factually incorrect or nonsensical outputs, particularly in technical and domain-specific settings \citep{rawte2023survey, ji2023towards, huang2025survey}. This issue has motivated substantial research efforts both in understanding the underlying causes of hallucination \citep{huang2025survey, mckenna2023sources} and in developing mitigation strategies against its occurrence \citep{ji2023towards}.
In scientific and engineering domains with significant real-world implications, such as nuclear physics and nuclear engineering, correct and reliable responses often require not only factual recall but also precise quantitative reasoning and a clear understanding of the domain concepts. Errors in these contexts, whether due to hallucination, incorrect application of formulae and calculations, or shallow semantic understanding, can significantly limit the practical utility of large language models. In short, when it comes to technical topics, there is significant risk if LLMs fail to provide accurate and verifiable information.

To validate the performance and reliability of large language models, researchers have developed benchmarks \citep{liang2022holistic, chang2024survey}. Benchmarks typically consist of tasks designed to evaluate specific aspects of model behavior, each aiming to target a certain aspect of the response behavior of the LLMs, including complex contextual understanding, multi-step logical reasoning, and quantitative calculation \citep{hendrycks2020measuring, srivastava2023beyond}. 
Such benchmarks enable systematic assessment of core LLM capabilities.
However, a central challenge in assessing LLM performance in highly technical and specialized domains has been the lack of high-quality, scalable benchmarks. While many existing benchmarks already evaluate general reasoning and domain knowledge, they often fall short in capturing the unique approaches required in technical practice \citep{huang2025survey}. Some focus primarily on mathematical problem solving without domain specific settings \citep{cobbe2021training}, while others rely on small, expert-curated datasets that are costly to construct and difficult to scale \citep{singh2025ubiquity}. In the nuclear engineering domain, there has been limited prior work on constructing a benchmark for evaluation \citep{acharya2023nuclearqa}. It provides an important starting point, but is limited in size and requires substantial human effort for both dataset creation and evaluation.

This highlights the need for scalable, domain-specific benchmarks that can capture both conceptual understanding and quantitative reasoning in highly technical fields. In this work, we introduce NuclearQAv2, a new benchmark designed to evaluate LLMs on nuclear engineering knowledge through a structured and scalable approach. The benchmark consists of tasks in the form of question and answer (QA) pairs to assess the ability of LLMs to understand the domain knowledge and carry out related technical execution. A key aspect of our design is the use of multiple question types to capture different facets of domain competence of language models. Specifically, the benchmark consists of three categories of question answering tasks: (i) boolean (yes/no) questions that assess factual grounding of the subject, (ii) numeric questions that require more complex multi-step logical deduction, calculation, and quantitative reasoning, and (iii) verbal questions that test conceptual understanding through short-form answers. The three categories are designed to test different capabilities separately, and collectively provide a more comprehensive evaluation of model capabilities than single-format benchmarks.

To enable scalable benchmark construction, we have developed a hybrid pipeline that combines human-authored questions, existing data, and automatically generated question-answer pairs derived from domain-specific textual corpora. We leverage document parsing tools to extract structured text from technical materials and employ large language models to generate candidate QA pairs under carefully designed prompting constraints. This approach allows us to efficiently expand dataset coverage while maintaining control over question quality and relevance. The resulting NuclearQAv2 dataset contains approximately 1,240 QA pairs spanning the three question types.
On the other hand, efficient evaluation of model outputs across these heterogeneous tasks introduces additional challenges. While boolean and numeric questions can be assessed using straightforward accuracy metrics (with tolerance for numerical variation), verbal responses require handling linguistic variability such as synonyms and paraphrases. To address this, we adopt an LLM-based evaluation scheme for verbal questions, using a structured prompting approach to approximate semantic equivalence between generated and referenced answers. This enables automatic evaluation of model responses and computation of benchmark metrics. 
We evaluate several popular large language models against NuclearQAv2 and observe their performance across question types. In particular, language models tend to achieve higher accuracy on boolean questions, while performance on the complex numeric tasks remains substantially lower, and verbal question performance varies depending on the model. These results highlight the importance of multi-faceted evaluation in understanding the strengths and limitations of LLMs in nuclear engineering.

Our main contributions in this paper are threefold: (i) We introduce NuclearQAv2, a novel benchmark dataset for evaluating LLMs on nuclear engineering knowledge, comprising $\sim 1,240$ QA pairs across three distinct question types. The benchmark captures factual, quantitative, and conceptual aspects of domain-science competence, enabling systematic evaluation of this highly technical domain. (ii) We develop a hybrid benchmark dataset construction pipeline that combines human expertise and LLM-based generation from technical corpora, enabling scalable benchmark development while reducing manual effort. (iii) We introduce an automated evaluation framework that supports both objective and free-form responses through task-specific metrics and LLM-based semantic assessment.
Overall, NuclearQAv2 provides a scalable and extensible benchmark for assessing LLM performance in nuclear engineering, and offers insights into the capabilities and limitations of current models in domain-science reasoning.

\section{Related Works}
\label{sec:lit}

A growing body of works have explored the evaluation of LLMs in specialized subjects using domain-specific benchmarks. These benchmarks aim to assess domain knowledge beyond general-purpose datasets, often relying on expert-curated question–answer pairs and tasks derived from real-world practice. Notable examples include medical QA benchmarks such as MedQA and PubMedQA \citep{singhal2023large}, legal reasoning benchmarks such as LegalBench \citep{guha2023legalbench}, and expert-authored datasets on professional reasoning such as ExpertQA \citep{malaviya2024expertqa} and PRBench \citep{akyurek2025prbench}, which incorporate tasks and evaluation criteria designed by domain professionals.
In the nuclear engineering domain, the closest work to NuclearQAv2 is the NuclearQA benchmark developed in \cite{acharya2023nuclearqa}. This work introduced a benchmark consisting of 100 QA pairs crafted by human experts on nuclear topics, which cover multiple question types. It features both short-answer questions that primarily assess factual knowledge, and open-ended long-answer questions, which require additional reasoning abilities. The questions are split into different types, including numeric and scientific, or a combination of both. The evaluation of this benchmark requires human input, and has a five-tier scorecard for each task from correct to nonsensical responses. 
While such efforts produce benchmark data of high quality, they are typically small in scale and expensive to construct, limiting their coverage and extensibility.
This is indeed the case for the NuclearQA benchmark. The requirement for expert involvement during both benchmark construction and evaluation limits the scalability and reproducibility of the benchmarking process.

Recent works have also explored the use of LLMs for automated dataset construction, including the generation of synthetic QA pairs from textual corpora \citep{alberti2019synthetic,puri2020training}. These approaches enable rapid scaling of datasets and reduce reliance on costly human annotation, but introduce challenges related to quality control, diversity, and faithfulness to source material. In practice, the quality of generated QA pairs can vary significantly, necessitating additional filtering or selection mechanisms to ensure usefulness \citep{jin2024select}. 
Prior methods often rely on prompt engineering to guide generation, but may lack systematic constraints to ensure that generated questions are non-trivial, unambiguous, and answerable without additional context \citep{alberti2019synthetic}.

Expert-authored benchmarks provide high-quality evaluation data but are costly to expand, whereas automatically generated datasets offer scalability at the expense of potential quality degradation. These trade-offs motivate hybrid benchmark construction approaches that combine expert knowledge with automated generation.

Beyond dataset generation for benchmarks, significant efforts have been made on evaluating LLMs on quantitative and multi-step reasoning tasks, including mathematical problem-solving benchmarks such as GSM8K and MATH \citep{cobbe2021training, hendrycks2021measuring}. These datasets have been instrumental in measuring progress in arithmetic and symbolic reasoning. However, they are generally domain-agnostic, emphasizing abstract problem solving rather than knowledge grounded in a specific scientific discipline \citep{liang2022holistic}.
In contrast, real-world technical tasks often require a combination of factual recall, numerical computation, and conceptual understanding. Existing benchmarks rarely capture this combination within a single evaluation framework \citep{liang2022holistic, huang2025survey}. 
This leaves an important gap in the assessment of technical domain competence.
In addition to task design, evaluating free-form or short-answer responses remains a significant challenge, particularly in settings where exact string matching is insufficient due to paraphrasing or synonymy \citep{rajpurkar2016squad,zhang2019bertscore}. To address this, recent works have explored the use of large language models as automated evaluators, leveraging their ability to assess semantic similarity and overall response quality \citep{liu2023g}. While this approach enables scalable and flexible evaluation, it is sensitive to prompt design, model bias, and variability in judgment \citep{zheng2023judging}.
These observations motivate the development of evaluation frameworks that combine automated assessment with task-specific scoring mechanisms to support scalable benchmarking.

\section{The NuclearQAv2 Benchmark}
\label{sec:benchmark}
In this section, we first define the question answering tasks for the proposed NuclearQAv2 benchmark, and we introduce the three types of tasks/questions in detail. 
Subsequently, we discuss the design principles for building the benchmark, ensuring its capacity as a multi-faceted evaluation framework for domain science tasks.

\subsection{Task Definition}
We consider the task of domain-specific question answering (QA) in nuclear engineering. Given a question denoted by $q$, a language model to be evaluated is required to produce a predicted answer, denoted by $\hat{a}$, consistent with domain knowledge. Depending on the question type, the expected answer (denoted by $a$) may take different forms, including a binary label/choice, a numerical value, or a short textual response.

Formally, the benchmark consists of a set of tasks each including three elements:
\begin{equation}
    \mathcal{D} = \{ (q_i,a_i,t_i) \}_{i=1}^N,
\end{equation}
where $q_i$ is the question, $a_i$ is the reference answer, and $t_i\in\{\text{boolean, numeric, verbal}\}$ denotes the question type. Each question type is associated with a specific evaluation protocol, as described in Section~\ref{sec:eval}. 

\subsection{Question Type Taxonomy}
NuclearQAv2 utilizes three distinct question types, each targeting a different aspect of domain-science competence. We provide examples for each type to showcase the technical nature of the questions in nuclear engineering.

\paragraph{Boolean Questions (Factual Grounding)}
The first category of questions are boolean questions which require a binary response (e.g., ``yes'' or ``no'', ``true'' or ``false'') and are designed to assess factual knowledge. These questions typically involve well-defined statements where correctness can be unambiguously determined automatically. This type of questions focus on checking fundamental domain knowledge in a clear manner, enabling efficient evaluation.
While boolean questions are easy to evaluate, there is a risk that language models may guess or exploit response-pattern biases. Therefore, this type of questions primarily test whether the model can establish correct factual associations.

\noindent\textbf{Example}:\\
\texttt{Question: True or False: the decay energy spectra for beta and \\
alpha decay are both continuous.}\\
\texttt{Expected answer: False}

\paragraph{Numeric Questions (Quantitative Reasoning)}
Numeric questions require the language model to produce a number as the answer, involving application of the correct formulae in nuclear engineering and possibly multi-step calculation. These questions are designed to assess quantitative reasoning and logical process when facing potentially complex tasks requiring multiple steps of deductions.
For numeric tasks, we use a tolerance-based evaluation criterion, only allowing small deviations in the results caused by rounding calculations and using slightly different constants or coefficients. In general, the numeric questions provide a stronger signal of whether the model can perform domain-relevant calculations rather than relying on surface-level patterns.

\noindent\textbf{Example}:\\
\texttt{Question: How much of natural uranium feed is required to produce\\
enriched fuel containing 1kg of $U^{235}$? Provide the estimate in unit\\
of kg. Answer the question only with a number without unit.\\
Provide the numeric result only without reasoning.}\\
\texttt{Expected answer: 200}

\paragraph{Verbal Questions (Conceptual Understanding)}
The third type of tasks are the verbal questions. They require short textual answers, typically consisting of a few words or phrases. These questions assess conceptual understanding and semantic knowledge within the domain.
Evaluation of verbal responses is inherently more challenging due to linguistic variability, including synonyms and variations in phrasing. However, they are important for capturing whether the model can correctly identify and articulate domain-specific concepts. 

\noindent\textbf{Example}:\\
\texttt{Question: What do you call nuclei which contain the same number of\\
protons but different numbers of neutrons? Answer the question only\\
with several words without punctuation.}\\
\texttt{Expected answer: isotopes}

\subsection{Design Principles}
\label{sec:principle}
To ensure the quality and usefulness of the benchmark, and to facilitate the evaluation process, we adopt several principles when constructing the QA pairs for NuclearQAv2. 

We follow five principles when constructing questions: (i) Each question belongs to exactly one of the three types (boolean, numeric, or verbal). This is to prevent ambiguity in both task formulation and the expected answer for evaluation. (ii) The questions are designed to explicitly require nuclear engineering knowledge or reasoning, and to avoid becoming overly simple or surface-level. (iii) Each question is formulated such that there exists an unambiguous and well-defined answer, minimizing dependence on external context. This is especially crucial for verbal questions, where targeted answers need to be clearly defined instead of generic. (iv) Expected answers need to be concise. This enables automatic evaluation without human intervention. (v) Finally, the questions should be constructed to be context independent. This means they can be answered without access to the original source text.
These principles are enforced during both manual curation of QA pairs and automated generation conducted using LLMs in Section~\ref{sec:auto}, through prompt design and filtering criteria. This facilitates the subsequent evaluation stage as well.

The combination of boolean, numeric, and verbal question types provides a multi-faceted evaluation framework for domain-science QA. Rather than treating understanding the domain knowledge as a single capability, this structure enables more granular analysis of model performance across different reasoning modes.
In particular, boolean questions primarily test factual recall, numeric questions emphasize quantitative reasoning, and verbal questions assess conceptual understanding. This separation allows us to identify systematic differences in model behavior that may not be visible in single-format benchmarks.

\section{Dataset Construction}
\label{sec:const}
With the types of tasks specified, and the design principles established, we construct the NuclearQAv2 benchmark using a hybrid pipeline that combines human-authored questions, existing questions with answers, and automatically generated QA pairs derived from nuclear engineering textual corpora. In this work, we use the nuclear engineering textbook as the source document to build all the QA pairs \citep{lamarsh1977introduction}.
This approach aims to balance data quality and scalability, addressing the limitations of purely manual or fully automated dataset construction. 

We construct the dataset using the textbook with assistance from domain experts. For part of the boolean questions, we obtain QA pairs from domain experts within the coverage of the textbook. The questions focus on factual statements directly connected to the information of nuclear engineering. This ensures high-quality coverage of core nuclear engineering concepts.
For the numeric questions, we look into the exercise quantitative questions provided by the authors of the textbook. This helps ensure that the numeric questions cover a range of difficulty and complexity levels, while maintaining the quality of the QA pairs constructed.
For the verbal questions and some of the boolean questions, we directly look into the technical information and terminology provided in the main chapters of the textbook itself. To build the benchmark efficiently, we utilize LLM to carry out automated QA generation for these two types of questions. 

To enable automated processing and task generation, we convert the source document from PDF format into structured text using document parsing tools called Nougat \citep{blecher2023nougat}. This allows both natural language content and mathematical expressions to be extracted for downstream use in QA generation. For simplicity, we exclude the image information of the textbook from the generation process. 

\subsection{Automated QA Generation}
\label{sec:auto}
To construct the boolean and the verbal questions efficiently, we employ LLMs to generate the QA pairs automatically from the parsed text corpus. For NuclearQAv2, we utilize the Meta Llama 3 70B Instruct model for this purpose.
Given a portion of technical text, the LLM is prompted to produce multiple question–answer pairs that satisfy the set of constraints and principles established in Section~\ref{sec:principle} designed to ensure quality and relevance.
If these criteria cannot be satisfied for a given text passage, the generation process discards the created questions. This helps reduce noise and prevents the inclusion of low-quality or ill-posed questions.

We also recognize the quality of automatically generated QA pairs is highly dependent on prompt design. To guide the generation process, we employ structured prompts that encode these principles, including instructions to avoid ambiguous or context-dependent questions, exclude trivial queries (for example, questions about document structure and location), and ensure answers are concise and directly derived from domain knowledge. The prompt also provides instructions on the form of the answer being sought.
The prompt encourages diversity in question types while maintaining adherence to the predefined taxonomy (boolean or verbal). By explicitly specifying these constraints in prompts provided to the LLM, we improve the consistency and usability of generated QA pairs.

Following generation, QA pairs undergo a filtering and curation process to improve dataset quality. This includes removal of invalid outputs including incomplete QA pairs, and screening for ambiguity to remove questions which are unclear and may lead to multiple plausible answers. This step is currently carried out manually to ensure the quality of the QA pairs generated.

\subsection{NuclearQAv2 Dataset}
The final dataset consists of 1,239 QA pairs, split across three question types as follows:
\begin{itemize}
    \item Boolean (Yes/No): 750 questions;
    \item Numeric: 206 questions;
    \item Verbal: 283 questions.
\end{itemize}
This split reflects a balance between different reasoning modes. Boolean questions provide broad coverage of factual knowledge, while numeric and verbal questions target more complex reasoning and conceptual understanding.

The hybrid construction pipeline enables scalable dataset creation while maintaining control over question quality. By combining human expertise with LLM-assisted generation and structured filtering, NuclearQAv2 achieves broader coverage than purely manual benchmarks, while mitigating common risks associated with fully automated approaches.
However, the quality of the dataset remains influenced by the underlying generation model and prompt design. We discuss these limitations further in Section~\ref{sec:exp}.

\section{Evaluation of NuclearQAv2}
\label{sec:eval}
Evaluating LLM performance on NuclearQAv2 requires handling heterogeneous answer types, including binary labels, numerical values, and short textual responses. To ensure consistency and reproducibility, we define evaluation metrics for each question category based on accuracy.
Given a model response $\hat{a}$ and the corresponding reference answer $a$, correctness is determined according to the question type $t\in\text{\{boolean,numeric,verbal}\}$. For each type of tasks, we compute accuracy as the proportion of correctly answered questions. Overall performance is reported both per category and as an aggregate across all question types, weighted using the number of questions from each category.

For boolean questions, the model is required to output a binary response (e.g., true/false or yes/no). Evaluation is performed using exact match accuracy, where a prediction is considered correct if it matches the reference label.
To account for formatting variation, responses are normalized (e.g., mapping `yes' and `true' to a common label) prior to evaluation. This ensures that equivalent answers are treated consistently.

Numeric questions require the model to produce a numerical value with specified units, derived through calculation or unit conversion. Because minor differences in coefficients or rounding may occur, we evaluate predictions using a tolerance-based accuracy metric. A prediction $\hat{a}$ is considered correct if it satisfies:
\begin{equation}
    \frac{|\hat{a}-a|}{|a|}\leq \epsilon,
\end{equation}
where $\epsilon$ is a preset tolerance. Here we set $\epsilon=0.15$. This approach allows for small numerical deviations while still enforcing correctness in the underlying reasoning process.

Verbal questions require short textual answers, typically consisting of domain-specific terms or phrases. Exact string matching is insufficient in this setting due to linguistic variability, including synonyms, paraphrases, and minor differences in phrasing.
To address this, we employ an LLM-based evaluation scheme, in which a separate language model acts as a `judge' to determine whether a generated answer is semantically equivalent to the reference answer. The judge is prompted with both the generated and reference answers and instructed to return a binary decision (correct or incorrect) based on semantic similarity.
The evaluation prompt is designed to facilitate identifying equivalent wording or partial lexical overlap, penalize incorrect or unrelated answers, and maintain consistency across different phrasing variations.
This approach enables scalable evaluation of conceptual questions while approximating human judgment. The prompt utilized for the LLM judge is as follows:\\

\noindent\texttt{prompt: Act as an impartial grader to the answer of a question in \\
nuclear engineering. The guidance is that as long as the attempted \\
answer has the same meaning it is good. It is not necessary to obtain\\
the exact wording to be considered correct, but a correct answer\\
should at least partially share the same wording with the true answer.\\
The attempted answer is \emph{\{evaluated model answer\}}.\\
The true answer is \emph{\{reference answer\}}.\\
Provide the response only with either `correct' or `incorrect'.}\\

With the procedures set up, the accuracy can be obtained from the response of the LLM judge for each task.
To ensure consistency, we use a fixed evaluation prompt and a single judge model across all verbal questions. All evaluation procedures are deterministic given model outputs and predefined parameters, enabling reproducibility.

While the evaluation protocol is designed to be practical and scalable, we recognize that it still has several limitations. The accuracy of verbal evaluation depends on the reliability and consistency of the LLM judge itself, which may introduce bias or variability. The choice of tolerance $\epsilon$ may affect reported performance, particularly for questions requiring high precision. Finally, the use of binary accuracy, calculated from either correct or incorrect response, as the primary metric does not well capture partial correctness or intermediate reasoning quality for complex and multi-step tasks.

\section{Experiments}
\label{sec:exp}

We evaluate the performance of large language models on NuclearQAv2 using the prompting and evaluation protocol introduced in Section~\ref{sec:eval}. 
For each question, the model is prompted to produce a direct answer consistent with the expected output format (binary label, numerical value, or short phrase). We then report accuracy as the performance metric, computed separately for each of the three question types, and then aggregated together for a final score.

We evaluate nine popular language models of different scales with the NuclearQAv2 benchmark: (i) OpenAI GPT-5.2, (ii) OpenAI GPT-5.4, (iii) Meta Llama 3 70B Instruct, (iv) Amazon Nova Pro, (v) Mistral 7B Instruct, (vi) OpenAI GPT-4o, (vii) OpenAI gpt-oss-120B, (viii) NVIDIA Nemotron-3 Super 120B A12B FP8, and (ix) Meta Llama 3 8B Instruct. The results are summarized in Table~\ref{tbl:res}. The best accuracy numbers for each task are marked in red. The evaluation is carried out three times for each LLM to estimate mean accuracy and standard error.

\begin{table}[htbp]
\caption{Evaluation results for NuclearQAv2}
\label{tbl:res}
\vspace{0.5em}
\centering
\resizebox{\linewidth}{!}{%
\begin{tabular}{c|ccc}
\textbf{Language Model}               & \textbf{Boolean Tasks} & \textbf{Numeric Tasks} & \textbf{Verbal Tasks} \\ \hline\hline
OpenAI GPT-5.2                        & \textcolor{red}{0.8022 ($\pm$ 0.0043)}          &  0.6651 ($\pm$ 0.0074)         & \textcolor{red}{0.8233 ($\pm$ 0.0020)}\\
OpenAI GPT-5.4                        & 0.7991 ($\pm$ 0.0012)          &  0.7233 ($\pm$ 0.0148)         & 0.8104 ($\pm$ 0.0031)         \\
Meta Llama 3 70B Instruct             & 0.7600 ($\pm$ 0.0008)          & 0.5712 ($\pm$ 0.0016)          & 0.6749 ($\pm$ 0.0020)         \\
Amazon Nova Pro                       & 0.5698 ($\pm$ 0.0029)          & 0.5841 ($\pm$ 0.0098)          & 0.7079 ($\pm$ 0.0047)         \\
Mistral 7B Instruct                   & 0.3831 ($\pm$ 0.0022)          & 0.0307 ($\pm$ 0.0016)          & 0.5159 ($\pm$ 0.0020)         \\
OpenAI GPT-4o                         & 0.7307 ($\pm$ 0.0055)          & 0.6699 ($\pm$ 0.0028)          & 0.7479 ($\pm$ 0.0012)         \\
OpenAI gpt-oss-120B                   & 0.7960 ($\pm$ 0.0008)          & \textcolor{red}{0.7913 ($\pm$ 0.0048)}     & 0.7833 ($\pm$ 0.0072)         \\
NVIDIA Nemotron-3 Super 120B & 0.7924 ($\pm$ 0.0031)          &  0.7411 ($\pm$ 0.0016)                 & 0.7927 ($\pm$ 0.0083)         \\
Meta Llama 3 8B Instruct              & 0.7173 ($\pm$ 0.0000)          & 0.4709 ($\pm$ 0.0000)          & 0.4865 ($\pm$ 0.0031)        
\end{tabular}
}
\end{table}

Overall, the results demonstrate clear variation in LLM performance across the three types of tasks, highlighting the importance of evaluating factual, quantitative, and conceptual capabilities separately rather than only relying on a single aggregate metric. 

For the boolean tasks, which primarily assess factual knowledge and grounding in nuclear engineering concepts, the strongest performance is achieved by OpenAI GPT-5.2 with an average accuracy of $0.8022$. It is closely followed by GPT-5.4, OpenAI gpt-oss-120B, and NVIDIA Nemotron-3 Super 120B, with accuracy just short of $0.8$. The relatively small performance differences among the leading models suggest that factual recall and recognition-based reasoning are comparatively mature capabilities for current frontier LLMs. In contrast, smaller models such as Mistral 7B Instruct and Meta Llama 3 8B Instruct exhibit noticeably lower performance, indicating that factual domain knowledge remains sensitive to model scale and training quality.

The numeric tasks prove substantially more challenging as expected. These questions require multi-step reasoning, numerical computation, and correct application of formulae in nuclear engineering, making them less reliant on memorized knowledge alone. 
This category also exhibits the largest performance variation across models.
OpenAI gpt-oss-120B achieves the highest observed accuracy in this category with an average accuracy of $0.7913$. NVIDIA Nemotron-3 Super 120B and GPT-5.4 also demonstrate strong quantitative capabilities, whereas several models experience significant degradation relative to their boolean-task performance. The most pronounced example is Mistral 7B Instruct, whose mean accuracy drops to $0.0307$, suggesting substantial difficulty in performing the multi-step quantitative reasoning required by the benchmark. More broadly, the wider performance spread observed in the numeric category indicates that quantitative reasoning remains a major differentiating factor among large language models.

For the verbal tasks, which assess conceptual understanding through short-form free-text responses, OpenAI GPT-5.2 again achieves the highest average score of $0.8233$, followed closely by GPT-5.4, NVIDIA Nemotron-3 Super 120B, and OpenAI gpt-oss-120B. The generally strong performance of these leading models suggests that they are capable of capturing and expressing many core concepts in nuclear engineering. Nevertheless, the performance gap between frontier models and smaller open-weight models remains considerable, indicating that conceptual understanding in highly technical domains continues to benefit from increased model capability and scale. 

With three types of tasks combined, the aggregate accuracy scores are summarized in Table~\ref{tbl:res-agg}. We observe that the gpt-oss-120B model achieves the highest aggregated accuracy, where the GPT-5.2, GPT-5.4, and the Nemotron-3 Super 120B models follow closely.

\begin{table}[htbp]
\caption{Aggregated evaluation results for NuclearQAv2}
\vspace{0.5em}
\label{tbl:res-agg}
\centering
\resizebox{0.7\linewidth}{!}{%
\begin{tabular}{c|c}
\textbf{Language Model}               & \textbf{Aggregated Accuracy} \\ \hline\hline
OpenAI GPT-5.2                        & 0.7842 ($\pm$ 0.0034)       \\
OpenAI GPT-5.4                        & 0.7891 ($\pm$ 0.0033)       \\
Meta Llama 3 70B Instruct             & 0.7092 ($\pm$ 0.0007)       \\
Amazon Nova Pro                       & 0.6037 ($\pm$ 0.0038)       \\
Mistral 7B Instruct                   & 0.3548 ($\pm$ 0.0007)       \\
OpenAI GPT-4o                         & 0.7245 ($\pm$ 0.0031)       \\
OpenAI gpt-oss-120B                   & \textcolor{red}{0.7923 ($\pm$ 0.0015)}       \\
NVIDIA Nemotron-3 Super 120B A12B FP8 & 0.7840 ($\pm$ 0.0007)       \\
Meta Llama 3 8B Instruct              & 0.6236 ($\pm$ 0.0007)       
\end{tabular}
}
\end{table}

We obtain several observations from these results. (i) Most models achieve their highest performance on boolean tasks, reflecting the relative ease of factual recognition compared to open-ended reasoning and calculation. (ii) Numeric tasks consistently remain the most challenging category for many models, reinforcing prior observations that strong language understanding does not necessarily translate to reliable quantitative reasoning. (iii) The differences in performance across task categories demonstrate the value of a multi-faceted benchmark design. Models that perform strongly on factual questions do not always exhibit equivalent strengths in numerical reasoning or conceptual explanation, suggesting that a comprehensive evaluation of domain-science competence indeed requires assessment across multiple dimensions.


The results on NuclearQAv2 evaluation highlight some important considerations for evaluating large language models in technical domains. First, the observed performance differences across question types suggest that domain-science competence is not uniform, but instead depends strongly on the type of reasoning evaluated. In particular, we see some LLMs demonstrate strong performance on boolean questions, which primarily assess factual recall, but exhibit substantially lower accuracy on numeric and verbal tasks that require more complex quantitative reasoning and conceptual understanding.
These findings underscore the importance of multi-type evaluation frameworks. Benchmarks that rely on a single question format may provide an incomplete or overly optimistic view of model capabilities. By explicitly separating factual, quantitative, and conceptual components, NuclearQAv2 enables more fine-grained analysis of model strengths and weaknesses.
The results also suggest that model scale alone does not appear to fully resolve limitations in domain-specific reasoning. While larger models show consistent improvements across all categories, the performance gap between boolean and numeric questions persists. This indicates that additional measures, such as improved training data, domain-specific fine-tuning, or integration with external tools, may become necessary to achieve reliable performance in technical settings.

\section{Discussion}
In this work, we have introduced NuclearQAv2, a benchmark for evaluating large language models on nuclear engineering knowledge through a combination of factual, quantitative, and conceptual question-answering tasks. The benchmark comprises 1,239 QA pairs covering three distinct task categories, including boolean, numeric, and verbal, which are designed to assess complementary aspects of domain-science competence.
To support scalable benchmark development, we have proposed and implemented a hybrid construction pipeline that combines expert-authored questions, existing benchmark data, and LLM-assisted generation from domain-specific technical corpora. We further developed a structured automated evaluation framework with task-specific assessment methods, including tolerance-based scoring for numeric responses and LLM-based semantic evaluation for verbal answers.

Experimental results across several LLMs reveal substantial differences in performance across task types. While many models demonstrate strong performance on factual questions, quantitative reasoning and conceptual understanding remain considerably more challenging. These findings suggest that performance on a single task format is insufficient for characterizing model capabilities in technical domains and highlight the importance of multi-faceted evaluation frameworks as proposed in this work. 
Furthermore, the observed performance differences among the evaluated models indicate that NuclearQAv2 is capable of distinguishing meaningful variations in nuclear engineering tasks and therefore serves as an effective benchmark for assessing language models.

Several limitations of the current work motivate future research. The prompts used for automatic generation of QA pairs for verbal tasks have been engineered based on manual assessment of quality. 
Future work will investigate more systematic approaches to prompt optimization and quality assurance.
Additionally, the automatic generation currently relies on an older language model, which may limit the quality of the questions generated. On the other hand, the LLM-based evaluation of verbal responses may be sensitive to evaluator choice and prompt formulation, and future work should compare automated judgments against expert human assessment.
Finally, the benchmark currently only handles textual information. Many scientific and engineering disciplines rely heavily on multi-modal information such as diagrams, figures, schematics, and plots. Extending the benchmark to incorporate multi-modal data represents an important direction for future development.

Overall, NuclearQAv2 provides both a practical benchmark for assessing LLM performance in nuclear engineering and a scalable methodology for constructing domain-specific evaluation datasets. This work has also provided a framework for more efficient and rigorous benchmark construction and automated LLM evaluation in scientific and engineering applications, contributing to the development of more reliable AI systems for high-impact technical domains.

\newpage
\bibliographystyle{plain}
\bibliography{references}

\newpage
\begin{center}

{\large\bf SUPPLEMENTARY MATERIAL}

\end{center}

\section{NuclearQAv2: Boolean Task Configuration Codes}
\begin{lstlisting}[language=ruby, breaklines=true, basicstyle=\small\ttfamily, showstringspaces=false, showspaces=false]
tag:
  - nuclear
task: nuclearqa

dataset_path: json
dataset_name: null

dataset_kwargs:
  data_files:
    dev: /home/nuclearqa_lm_eval/data/nuclearqa/dev.json
    validation: /home/nuclearqa_lm_eval/data/nuclearqa/validation.json
    test: /home/LLM/nuclearqa_lm_eval/data/nuclearqa/test.json

output_type: generate_until

generation_kwargs:
  do_sample: false
  until: []


training_split: dev
validation_split: validation
test_split: test

doc_to_text: | 
  Answer the question with a single LaTeX box on its own line. Do not include any explanation. Just provide the answer only.

  Question: {{question}}

  If the question asks for "True or False", respond with exactly one of \boxed{True} or \boxed{False}.

  If the question asks for "yes or no", respond with exactly one of \boxed{Yes} or \boxed{No}.
  
  Answer: 

doc_to_target: "{{answer}}"

metric_list:
  - metric: exact_match
    aggregation: mean
    higher_is_better: true
    ignore_case: true
    ignore_punctuation: true
    regexes_to_ignore:
      - "(?s)^.*?\\\\boxed\\{"   # strip up to and including \boxed{
      - "(?s)\\}.*$"             # strip closing } and anything after
      - "^\\s+"                  # trim leading whitespace
      - "\\s+$"                  # trim trailing whitespace

metadata:
  version: 2
  description: "A binary question-answering benchmark where models answer Yes or No/True of False. Updated to remove using log probs in evaluation."
\end{lstlisting}

\newpage
\section{NuclearQAv2: Numeric Task Configuration Codes}
\begin{lstlisting}[language=ruby, breaklines=true, basicstyle=\small\ttfamily, showstringspaces=false, showspaces=false]
tag:
  - nuclear
task: nuclearqanum

dataset_path: json
dataset_name: null

dataset_kwargs:
  data_files:
    dev: /data/dev_num.json
    validation: /data/validation_num.json
    test: /data/test_num.json

output_type: generate_until
generation_kwargs:
  until:
    - "\n"
    - "."
training_split: dev
validation_split: validation
test_split: test

doc_to_text: "Question: {{question}}?\nAnswer:" #!function utils.doc_to_text
doc_to_target: "answer"
metric_list:
  - metric: !function utils.numeric_accuracy
    aggregation: mean
    higher_is_better: true

metadata:
  version: 0
  description: "A numeric question-answering benchmark where models provides a number as answer."
\end{lstlisting}
\textbf{Utility functions}
\begin{lstlisting}[language=python, breaklines=true, basicstyle=\small\ttfamily, showstringspaces=false, showspaces=false]
def doc_to_text(doc):
    instruction = "Answer the question with a number without unit. Provide the numeric result only without reasoning."
    print(f"Question: {doc['question']}\n{instruction}\nAnswer:")
    return f"Question: {doc['question']}\n{instruction}\nAnswer:"

def numeric_accuracy(predictions, references):
    tolerance = 0.15  # 15% tolerance
    correct = 0
    total = len(predictions)
    print(predictions)
    print(references)
    for pred, ref in zip(predictions, references):
        try:
            pred_num = float(pred.strip())
            ref_num = float(ref.strip())
            if abs(pred_num - ref_num) / abs(ref_num) <= tolerance:
                correct += 1
        except:
            continue
    return correct / total #correct / total if total else 0
\end{lstlisting}

\newpage
\section{NuclearQAv2: Verbal Task Configuration Codes}
\begin{lstlisting}[language=ruby, breaklines=true, basicstyle=\small\ttfamily, showstringspaces=false, showspaces=false]
tag:
  - nuclear
task: nuclearqaverb

dataset_path: json
dataset_name: null

dataset_kwargs:
  data_files:
    dev: /data/nuclearqaverb/dev_verb.json
    validation: /data/nuclearqaverb/validation_verb.json
    test: /data/nuclearqaverb/test_verb.json

output_type: generate_until
generation_kwargs:
  until:
    - "\n"
    - "."
training_split: dev
validation_split: validation
test_split: test

doc_to_text: "Question: {{question}}?\nAnswer:"
doc_to_target: "answer"
metric_list:
  - metric: !function utils.verb_accuracy
    aggregation: mean
    higher_is_better: true

metadata:
  version: 0
  description: "A verbal question-answering benchmark where models provides several words as answer."
\end{lstlisting}
\textbf{Utility functions}
\begin{lstlisting}[language=python, breaklines=true, basicstyle=\small\ttfamily, showstringspaces=false, showspaces=false]
import json
from transformers import pipeline

def verb_valid(predictions, references):
    valid = 0
    total = len(predictions)
    for pred, ref in zip(predictions, references):
        if pred is not NA:
            valid += 1
    return valid / total if total else 0

def verb_accuracy(predictions, references):
    correct = 0
    total = len(predictions)
    pred_list = []
    ref_list = []
    for pred, ref in zip(predictions, references):
        if pred is not None:
            # Run a following step to determine whether the answer is correct
            classifier = pipeline("zero-shot-classification", model="EleutherAI/gpt-j-6B")
            prompt = f"Act as an impartial grader to the answer of a question in nuclear engineering. The guidance is that as long as the attempted answer has the same meaning it is good. It is not necessary to obtain the exact wording to be considered correct, but a correct answer should at least partially share the same is {pred}.\n The true answer is {ref}.\n Provide the response only with either 'correct' or 'incorrect'."
            result = classifier(prompt, candidate_labels=["correct", "incorrect"])
            label = result['labels'][0]
            if label == "correct":
                correct += 1
    
    return correct / total if total else 0
    
\end{lstlisting}

\end{document}